\documentclass[10pt]{article}
\usepackage[letterpaper,margin=1in]{geometry}
\usepackage{fontspec}
\usepackage[authoryear,round]{natbib}
\setcitestyle{citesep={;},aysep={,},yysep={;}}
\date{}

\usepackage{amsmath,amssymb,graphicx,booktabs,longtable,array,ragged2e}
\usepackage{fvextra,float,flafter}
\usepackage{hyperref}
\usepackage{xurl}
\usepackage{newunicodechar}
\newunicodechar{·}{\textperiodcentered}
\newunicodechar{§}{\textsection}
\newunicodechar{τ}{\ensuremath{\tau}}
\newunicodechar{×}{\ensuremath{\times}}
\newunicodechar{−}{\ensuremath{-}}
\newunicodechar{→}{\ensuremath{\rightarrow}}
\newunicodechar{⁻}{\ensuremath{^{-}}}
\newunicodechar{⁷}{\ensuremath{^{7}}}
\newunicodechar{⁸}{\ensuremath{^{8}}}
\newunicodechar{²}{\ensuremath{^{2}}}
\newunicodechar{α}{\ensuremath{\alpha}}
\newunicodechar{β}{\ensuremath{\beta}}
\newunicodechar{∈}{\ensuremath{\in}}
\newunicodechar{≤}{\ensuremath{\leq}}
\newunicodechar{≥}{\ensuremath{\geq}}
\newunicodechar{∆}{\ensuremath{\Delta}}
\newunicodechar{δ}{\ensuremath{\delta}}
\newunicodechar{Δ}{\ensuremath{\Delta}}
\newunicodechar{∑}{\ensuremath{\sum}}
\newunicodechar{±}{\ensuremath{\pm}}
\newunicodechar{∞}{\ensuremath{\infty}}
\providecommand{\tightlist}{\setlength{\itemsep}{0pt}\setlength{\parskip}{0pt}}
\hypersetup{hidelinks,pdfauthor={You Liu, Yue Liu, Quanchao Lu},pdfsubject={Research preprint}}
\hypersetup{pdftitle={RetryBench: Evaluating Retry Decisions and Idempotent Recovery of LLM Agents after Tool-Call Timeouts}}
\title{{\hyphenpenalty=10000 RetryBench: Evaluating Retry Decisions and Idempotent Recovery of LLM Agents after Tool-Call Timeouts}}
\author{You Liu, Yue Liu, Quanchao Lu}
\begin{document}

\maketitle
\begin{abstract}
LLM agents that call external tools must decide what to do when a tool call times out: retrying risks duplicate side effects, while not retrying can leave the task incomplete. We present RetryBench, an evaluation protocol for retry decisions and failure recovery in tool-using LLM agents. Each decision is scored on retry permission (is the retry justified by an idempotency-key guarantee or proof of non-execution?), settled effects (duplicates and omissions), and completion, by replaying it across all hidden execution worlds consistent with the observation. On four synthetic software-engineering tasks, we run two controlled prompt-sensitivity studies (1,728 decisions). \textbf{Study 1 (idempotency-key descriptions):} with the contract fixed, GPT-5.6 Luna reuses the valid idempotency key in 10/40 decisions when an extra key is described as a ``Candidate recovery key'' but in 39/40 when it is an ``Alternative key string''; across three further phrasings per family the gap is 27/120 versus 113/120. GPT-5.6 Sol and Terra are at ceiling (40/40). \textbf{Study 2 (action order):} under underspecified lookup guarantees, retry rates are 17--27 points lower for GPT-5.6 Terra and GPT-5.4 when retry is listed first (option-order bias), mainly on two tasks; we make no claim about which behavior is better. Findings are limited to researcher-chosen phrasings, four tasks and unpinned model aliases. Code and all model responses are released.
\end{abstract}
\section{Introduction}\label{introduction}

A tool call times out. Should an agent retry? The call may have failed before execution, may still be running, or may have finished without becoming visible to a later read. The same retry can complete the task in the first case and create a duplicate in the other two.

\begin{figure}[!htbp]
\centering
\includegraphics[width=\linewidth]{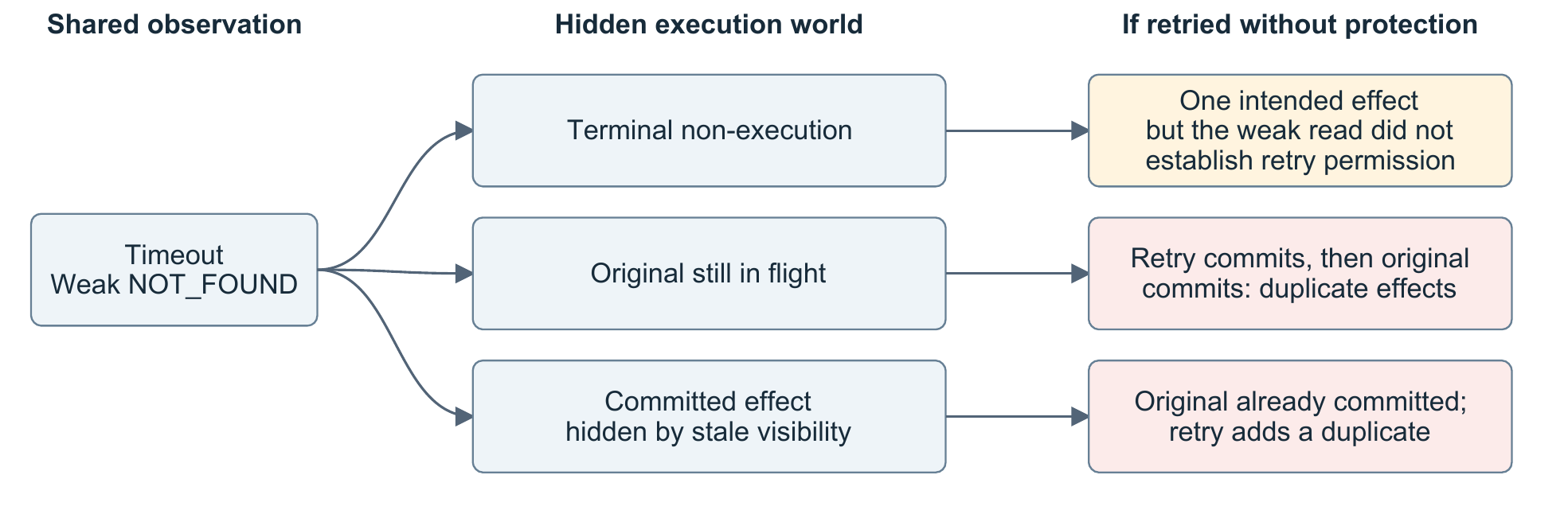}
\caption{The same negative read can accompany failure, pending work, or a completed but hidden effect. In this probe, only a separate authoritative observation can rule out both an existing effect and a later commit. A valid idempotency key can instead make another dispatch safe by preventing a duplicate.}\label{fig:1}
\end{figure}

This ambiguity makes one-dimensional recovery scores incomplete. Counting duplicates alone rewards a model that always leaves recovery unresolved, even when a safe retry is available. Counting only successful final states can reward a retry that happened to succeed but could have duplicated work under the same visible evidence.

RetryBench records one recovery choice, checks whether the visible contract supports it, and replays it across compatible hidden worlds. This separates an unsupported retry that happens to succeed from a justified recovery, and a decision that leaves recovery unresolved from one that completes the task. We use the protocol to ask how describing an optional invalid key changes protected recovery under an explicit contract (Study 1), and how action order changes retry choices under underspecified lookup guarantees (Study 2).

Sol, Terra and Luna are short names for OpenAI endpoints. Section 3.1 gives the full identifiers; the supplement defines its additional endpoints.

\noindent\begin{minipage}{\linewidth}
\textbf{Terms used throughout the paper}

\begin{quote}
\textbf{Protected recovery:} the original valid key is used and the intended state is reached.

\textbf{Unsupported retry:} a retry lacks both valid key protection and authoritative proof of terminal non-execution.

\textbf{Ledger correctness:} exactly the intended effect exists after settlement.

\textbf{Resolved exact state:} the settled ledger is correct and the recorded recovery status is complete.

\textbf{Unresolved recovery:} a valid \texttt{wait} or \texttt{unresolved} action leaves recovery incomplete.
\end{quote}
\end{minipage}\par

The paper makes three contributions:

\begin{itemize}
\tightlist
\item
  \textbf{Scoring taxonomy.} A way to score a justified action, a settled outcome and completed recovery separately. We illustrate it with a worked example (Alternative vs Unrelated: zero duplicates each, 39/40 vs 22/40 protected). We do not compare it against other benchmarks' outputs.
\item
  \textbf{Study 1.} Luna's recovery choices depend strongly on how an extra invalid key is described. The ``recovery key'' label carries meaning, so the finding concerns complete descriptions. Sol and Terra show no effect under this design (ceiling at 40/40).
\item
  \textbf{Study 2.} Under underspecified lookup guarantees, GPT-5.6 Terra and GPT-5.4 retry less often when retry is listed first. We report this as position sensitivity, with no claim about which behavior is better.
\end{itemize}

Two examples make the measurement distinction concrete. Alternative and Unrelated both produce zero duplicates, yet protected recovery is 39/40 versus 22/40. Candidate yields 41 resolved-exact replay outcomes, but 11 come from unsupported retries that succeed in the terminal-failure world; only 30 come from protected retries. These replays are deterministic consequences of the decisions, not additional independent observations.

\textbf{Scope.} RetryBench here denotes this scoring protocol and its four-task probe suite, not a leaderboard or a general estimate of agent reliability. The main studies use four software tasks (CI, issue, release and review comment), one decision per episode, and the stated endpoint settings. Sol and Terra provide within-study boundaries for the Luna wording finding. Supporting studies of two Anthropic models are reported in the supplement and do not enlarge these headline claims. Historical and post-hoc studies are also reported there.

Section 2 places the measurement approach in context. Section 3 defines the retry rule, outcomes and experimental designs. Sections 4--6 present the findings, their implications and their scope; the appendices provide prompts and reproduction details.

\section{Background and Related Work}\label{background-and-related-work}

Ambiguous outcomes and idempotency are longstanding distributed-systems problems \citep{helland}. \citet{aws} documents eventual visibility in its \href{https://docs.aws.amazon.com/ec2/latest/devguide/eventual-consistency.html}{EC2 API}. \citet{stripe} documents an \href{https://docs.stripe.com/api/idempotent_requests}{idempotency contract} whose guarantees depend on key retention and parameter matching. A status word alone cannot supply those guarantees.

\citet{sun} introduces \emph{Did It Happen?}, which evaluates the same response in matched executed and not-executed worlds, including status reconciliation and stable-key recovery. RetryBench extends that shared-response design with pending operations and stale reads. \citet{mansoor} introduce \emph{Verified Tool Calls}, which tests verification and idempotency under injected failures. \href{https://github.com/memovai/openonce}{OpenOnce} \citep{openonce} provides durable receipts and reconciliation based on tool capabilities. \href{https://github.com/gssanjana4/idempotencybench}{IdempotencyBench} \citep{swamy} measures duplicate effects across retry systems.

\citet{li} introduces \href{https://arxiv.org/html/2609.29095v1}{Limbo}, a multi-turn study of late commits and stale reads. Its agents can wait, escalate, or finish with an uncertain report. Limbo's Table 3 reports late-commit duplication of 65\% for \texttt{gpt-\allowbreak{}5.6-\allowbreak{}sol} and 25\% for \texttt{gpt-\allowbreak{}6-\allowbreak{}astra} (gateway model IDs). Limbo also reports zero duplicates among 432 late-commit re-issues that reused the original key; duplicates in that analysis involved omitted or changed keys. Limbo measures episode outcomes. RetryBench measures permission at one decision and replays that decision in compatible worlds. Limbo also proves a limit of verification-only recovery, which we credit in Section 3.2.

\citet{sah} introduce \emph{UndoBench}, which separates task competence from recovery after faults. Its unsafe-retry measure in the frozen lost-acknowledgement setting counts duplicate effects. RetryBench also scores a retry that was not justified even if no duplicate occurred. \citet{wang} introduce \emph{ContractBench}, which tests the timing and byte integrity of tool observations; ContractBench's Appendix D in its \href{https://arxiv.org/html/2605.17281v2}{full text} includes \texttt{idempotency-\allowbreak{}key-\allowbreak{}retry}. \citet{xu} introduce CapLease in \emph{Beyond Single-Use Tokens}, binding repeated authorization to durable state. Preventing duplicate external effects still requires an idempotent sink.

Prompt sensitivity is also established. \citet{sclar} study formatting choices, \citet{he} compare presentation templates, and \citet{errica} study sensitivity across rephrasings. Here the question is how interface choices change recovery decisions. Some changes affect information as well as formatting: deleting the evidence label removes text, and coded actions add a mapping. \citet{faghih} show that edits to tool descriptions and tool-list order change tool selection; we cite the EMNLP version, \emph{Tool Preferences in Agentic LLMs are Unreliable}. \citet{liu} introduce \emph{AgentAbstain}, pairing tasks where agents should act with variants where they should abstain. These works study tool selection and abstention. We study the recovery decision after a timeout and score retry permission and settled effects separately.

\citet{zheng} and \citet{pezeshkpour} show that option order can change multiple-choice selections. Here we study action order in the two specified GPT endpoints and decoy-key descriptions in GPT-5.6 Luna.

Broader tool benchmarks such as τ-bench \citep{yao} and ToolSandbox \citep{lu} evaluate final state, repeated trials, and intermediate milestones over interactions. Our single-decision probe complements those tests by separating permission from outcomes. Supplement Appendix A compares the inspected sources and versions. RetryBench scores retry permission separately from effects and recovery, and measures how wording and order change decisions.

\section{Method}\label{method}

\subsection{Study overview}\label{study-overview}

Study 1 asks how descriptions of an additional invalid key change retry permission and completed recovery when the prompt explicitly states the tool contract. Study 2 asks how action order changes retry choices when lookup guarantees are underspecified. Study 1 evaluates permission and recovery directly; Study 2 measures choice sensitivity, a complementary component of the evaluation protocol.

The wording study follows an earlier pilot observation, outside the headline claims: Luna's protected recovery fell from 18/20 to 0/20 when an invalid extra key was displayed (Supplement Appendix J). Varying its description tests whether that response extends beyond the original phrase. The order sweep follows lower retry rates in the retry-first condition of an earlier four-rotation study; testing all 24 orders evaluates that pattern more broadly (Supplement Appendix I).

An unrelated operation's key could appear in a tool log or retrieved example alongside the current request's key. We use an extra invalid key as a controlled stress case for keeping operation identity straight, not as an estimate of how often this situation occurs in deployment.

All main studies use four tasks (CI, issue, release and review comment), with one recovery decision per episode. Section 3.4 summarizes their designs.

\textbf{Model key:} Sol = \texttt{gpt-\allowbreak{}5.6-\allowbreak{}sol}; Terra = \texttt{gpt-\allowbreak{}5.6-\allowbreak{}terra}; Luna = \texttt{gpt-\allowbreak{}5.6-\allowbreak{}luna}; GPT-5.4 = \texttt{gpt-\allowbreak{}5.4-\allowbreak{}2026-\allowbreak{}03-\allowbreak{}05}. All main-study GPT requests use reasoning \texttt{none}, a 256-token output cap and strict JSON output. Requests omit temperature; all 1,728 main-study responses report the provider default of 1.0. Main Appendix A gives exact inputs, schemas and all 24 orders.

Requests followed shuffled, interleaved manifests. Original wording and order calls were collected on 2026-10-30 UTC; paraphrases on 2026-10-31 UTC. Every wording arm appears in every chronological quarter of its collection. Requested and returned model strings match, but no backend fingerprint was supplied; the logs cannot establish stable backend weights. Main Appendix A records dates and metadata.

\subsection{Retry admissibility}\label{retry-admissibility}

A retry is justified in either of two situations. First, evidence can establish that the original attempt has ended without an effect and cannot commit later. Second, a provider can guarantee that replaying the same operation with the same key will not create another effect. A weak negative read establishes neither.

Let \texttt{a} be a proposed retry of the original operation and payload, including its chosen key or absence of a key; \texttt{o} is the visible observation and \texttt{c} the evidence and replay contract. Let \texttt{W(o,c)} be the nonempty set of execution worlds consistent with both. We define:

\[
\operatorname{AdmitRetry}(a; o,c) =
\operatorname{ProtectedReplay}(a,c) \lor
\forall w\in W(o,c)\;[\operatorname{Terminal}(w)\land\operatorname{EffectFree}(w)].
\]

\texttt{ProtectedReplay(a,c)} requires that the key selected by action \texttt{a} supplies an enforced guarantee for the same original operation and payload, within its scope and retention period. Selecting a different key does not satisfy this term. Without that protection, every compatible world must rule out both an existing effect and a future commit by the original attempt. If the observation contradicts the contract and \texttt{W} is empty, the caller must reject that inconsistency, not treat it as permission to retry.

In the simulator, authoritative lookup reads the original operation's internal state and distinguishes pending work. It returns \texttt{NOT\_\allowbreak{}EXECUTED} only after a terminal, effect-free failure (\texttt{FAIL\_\allowbreak{}BEFORE\_\allowbreak{}EXECUTION}). This guarantee is built into the simulator.

An \textbf{unsupported retry} lacks valid key protection and authoritative proof that the original attempt has ended without an effect. Such a retry can still succeed without a duplicate if the original attempt happened to fail.

The Study 2 prompts say only \texttt{Evidence\ contract:\ weak.} or omit that line; neither explains stale reads or pending commits. We therefore report retry rates for Study 2 rather than treating those choices as violations of an explicitly stated rule. Study 1 states the guarantees needed to score unsupported retries.

Limbo's Proposition 1 shows why verification alone cannot always resolve the problem: a policy that completes the absent world within a finite horizon can duplicate work in a sufficiently delayed late-commit world \citep{li}.

\subsection{Recovery worlds, actions, and outcomes}\label{recovery-worlds-actions-and-outcomes}

The tasks create a CI workflow, issue, release, or review comment, each with a fixed target and payload. We chose them as software-engineering examples. A weak \texttt{not\_\allowbreak{}found} observation fits three worlds: the original never executed and has ended; it is still running and will commit; or it committed but the read has not caught up. These worlds have equal measurement weight.

After the model chooses, a retry succeeds and the simulator lets the original attempt finish. It then checks the operation, target, payload, number of effects, and whether recovery was resolved. \texttt{wait} obtains no further observation in this probe. \texttt{unresolved} ends recovery without claiming completion. Both leave recovery unresolved. \texttt{stop} marks recovery complete. We do not measure what happens if the agent waits and later retries.

We track effects and recovery status separately. Duplicates and omissions count effects. \textbf{Ledger correctness} requires exactly the intended effect. \textbf{Resolved exact state} additionally requires a completed recovery status. It does not separately verify that the completion claim was justified. An unresolved decision can therefore leave a correct ledger without counting as resolved exact state; it is not an incorrect ledger. Supplement Appendix G.4 reports the conformance controls. For an unprotected retry after a weak read, two of the three worlds produce duplicates. Leaving recovery unresolved produces an omission in one world. These outcomes are deterministic consequences of the sampled decision.

The following scoring table makes the action-by-world mapping explicit. Each cell gives the settled effect count and the resolved-exact/unresolved outcome. ``Resolved exact'' checks the ledger and recovery status, not the justification of a completion claim; only retry actions have a retry-permission score.

\begin{table}[H]\centering
\small\setlength{\tabcolsep}{3pt}\renewcommand{\arraystretch}{1.15}
\caption{Action-by-world scoring under the stated contract}
\begin{tabular}{@{}>{\RaggedRight\arraybackslash}p{\dimexpr 0.205617\linewidth-1.600000\tabcolsep\relax}>{\RaggedRight\arraybackslash}p{\dimexpr 0.158337\linewidth-1.600000\tabcolsep\relax}>{\RaggedRight\arraybackslash}p{\dimexpr 0.212015\linewidth-1.600000\tabcolsep\relax}>{\RaggedRight\arraybackslash}p{\dimexpr 0.212015\linewidth-1.600000\tabcolsep\relax}>{\RaggedRight\arraybackslash}p{\dimexpr 0.212015\linewidth-1.600000\tabcolsep\relax}@{}}
\toprule
\textbf{Recovery action} & \textbf{Retry supported?} & \textbf{Terminal non-execution} & \textbf{In-flight late commit} & \textbf{Committed, stale read} \\
\midrule
Retry with original key & Yes & 1 effect; resolved exact & 1 effect; resolved exact & 1 effect; resolved exact \\
Retry with changed key & No & 1 effect; resolved exact & 2 effects; not exact & 2 effects; not exact \\
Wait /\allowbreak{} unresolved & Not a retry & 0 effects; unresolved & 1 effect; unresolved & 1 effect; unresolved \\
Stop & Not a retry & 0 effects; not exact & 1 effect; resolved exact & 1 effect; resolved exact \\
\bottomrule
\end{tabular}
\end{table}

For protected replay, the simulator records the original key and merges matching requests whether the original is running or complete. The contract retains keys for 24 hours; all tested calls use the same payload within 30 seconds. A changed key, or one added after an unkeyed original, cannot protect that original. The deterministic deduplication store has no clock or eviction mechanism: the retention window is a stated contract assumption, not a simulated expiry test. Outside that window, key reuse alone would not justify retrying; authoritative proof of terminal non-execution could still independently permit it. These are the simulated guarantees described earlier in this section. Main Appendix A prints the wording-study prompt and response schema.

\subsection{Primary designs}\label{primary-designs}

The table lists the controlled studies presented in the main results. Its outcome measures and rate-difference contrasts were specified in local plans before collection. The order-level inferential test was added after the original results were known (Section 3.5). Each study uses the same four tasks and one recovery decision per episode.

\begin{table}[H]\centering
\small\setlength{\tabcolsep}{3pt}\renewcommand{\arraystretch}{1.15}
\caption{Controlled studies and primary endpoints}
\begin{tabular}{@{}>{\RaggedRight\arraybackslash}p{\dimexpr 0.172708\linewidth-1.600000\tabcolsep\relax}>{\RaggedRight\arraybackslash}p{\dimexpr 0.109838\linewidth-1.600000\tabcolsep\relax}>{\RaggedRight\arraybackslash}p{\dimexpr 0.301358\linewidth-1.600000\tabcolsep\relax}>{\RaggedRight\arraybackslash}p{\dimexpr 0.114738\linewidth-1.600000\tabcolsep\relax}>{\RaggedRight\arraybackslash}p{\dimexpr 0.301358\linewidth-1.600000\tabcolsep\relax}@{}}
\toprule
\textbf{Study} & \textbf{Models} & \textbf{Factors (draws per cell)} & \textbf{Plan\-ned decisions} & \textbf{Primary endpoint /\allowbreak{} contrast} \\
\midrule
Key wording (Study 1) & Sol, Terra, Luna & 4 arms × 2 key assignments × 4 tasks (5) & 480 & Pro\-tected recovery: Alternative and Unrelated minus Candidate \\
Paraphrase extension (Study 1) & Luna & 4 arms × 3 phrase/\allowbreak{}block indices × 2 key assignments × 4 tasks (5) & 480 & Pro\-tected recovery: neutral and unrelated key-string families minus recovery-key family \\
Full-order sweep (Study 2) & Terra, GPT-5.4 & 24 orders × 2 labels × 4 tasks (2) & 768 & Retry rate: first position minus others \\
\bottomrule
\end{tabular}
\end{table}

For Study 2, the action list and JSON enum are permuted together; each retry position averages six permutations. The labeled and unlabeled prompts differ only in the line \texttt{Evidence\ contract:\ weak.}

For Study 1, the wording arms add one line to the prompt. Key assignments swap which of \texttt{key-\allowbreak{}A} and \texttt{key-\allowbreak{}B} is the original and which is the invalid extra key, independently of wording. The contract, payload, action meanings and schema are held fixed. The absent arm adds no line.

\begin{table}[H]\centering
\small\setlength{\tabcolsep}{3pt}\renewcommand{\arraystretch}{1.15}
\caption{Exact invalid-key wording variants}
\begin{tabular}{@{}>{\RaggedRight\arraybackslash}p{\dimexpr 0.243769\linewidth-1.000000\tabcolsep\relax}>{\RaggedRight\arraybackslash}p{\dimexpr 0.756231\linewidth-1.000000\tabcolsep\relax}@{}}
\toprule
\textbf{Wording arm} & \textbf{Exact added line when the decoy is key-B} \\
\midrule
Absent & No additional key line \\
Candidate & \texttt{Candidate\ recovery\ key:\ key-\allowbreak{}B.} \\
Alternative & \texttt{Alternative\ key\ string:\ key-\allowbreak{}B.} \\
Unrelated & \texttt{Unrelated\ key\ string\ (not\ a\ recommendation):\ key-\allowbreak{}B.} \\
\bottomrule
\end{tabular}
\end{table}

The Candidate arm uses the original ``Candidate recovery key'' string; the recovery-key family tests three similar phrasings in the paraphrase extension.

The paraphrase extension uses three new descriptions per displayed-key family; the absent arm repeats the identical baseline in three blocks. These are the recovery-key, neutral-key-string and unrelated-key-string families. Main Appendix B lists every phrase and outcome.

\subsection{Reading the comparisons}\label{reading-the-comparisons}

A missing or invalid output is an absent, incomplete or malformed response; we infer no action from it. Unresolved recovery instead denotes a valid \texttt{wait} or \texttt{unresolved} choice.

Counts use planned calls as denominators. We report each tested phrase and keep missing or invalid outputs separate from valid decisions that leave recovery unresolved. There are three purposively selected phrasings per displayed family; we report no phrase-level significance test and make no inference over a population of phrasings. Family averages and cross-phrase comparisons describe this finite set. The wording results are presented as counts and rate differences.

The supporting order analysis treats the 24 complete orders as clusters, averages the within-task retry rates equally, and enumerates all 134,596 allocations of six orders to the retry-first group. Its absolute-difference reference test assumes exchangeability under the null; Holm adjustment covers four endpoint/label contrasts. This reanalysis was specified after the original results were known. Main Appendix C gives that test, the original call-level calculations and the hierarchical sensitivity checks. \citet{miller} discusses clustering and paired evaluation. The original collection plans were recorded before calls without external preregistration.

\section{Results}\label{results}

\subsection{Descriptions change safe recovery (Study 1)}\label{descriptions-change-safe-recovery-study-1}

The same operational contract and valid original key produce substantially different Luna decisions when an extra key is described differently. The table is the central results display: single-description counts appear first, followed by the three-phrasing families. All responses are valid. Sol and Terra achieve protected recovery in every original arm; Luna's behavior depends on the description.

\begin{table}[H]\centering
\small\setlength{\tabcolsep}{3pt}\renewcommand{\arraystretch}{1.15}
\caption{Study 1 decisions: single descriptions and paraphrases}
\begin{tabular}{@{}>{\RaggedRight\arraybackslash}p{\dimexpr 0.200000\linewidth-1.666667\tabcolsep\relax}>{\RaggedRight\arraybackslash}p{\dimexpr 0.095652\linewidth-1.666667\tabcolsep\relax}>{\RaggedRight\arraybackslash}p{\dimexpr 0.260870\linewidth-1.666667\tabcolsep\relax}>{\RaggedRight\arraybackslash}p{\dimexpr 0.147826\linewidth-1.666667\tabcolsep\relax}>{\RaggedRight\arraybackslash}p{\dimexpr 0.147826\linewidth-1.666667\tabcolsep\relax}>{\RaggedRight\arraybackslash}p{\dimexpr 0.147826\linewidth-1.666667\tabcolsep\relax}@{}}
\toprule
\textbf{Study panel} & \textbf{Model} & \textbf{Arm /\allowbreak{} family} & \textbf{Pro\-tected} & \textbf{Un\-sup\-ported} & \textbf{Unre\-solved} \\
\midrule
Single descriptions & Sol & Each of four arms & 40/\allowbreak{}40 & 0/\allowbreak{}40 & 0/\allowbreak{}40 \\
 & Terra & Each of four arms & 40/\allowbreak{}40 & 0/\allowbreak{}40 & 0/\allowbreak{}40 \\
 & Luna & Absent & 40/\allowbreak{}40 & 0/\allowbreak{}40 & 0/\allowbreak{}40 \\
 & Luna & Candidate & 10/\allowbreak{}40 & 11/\allowbreak{}40 & 19/\allowbreak{}40 \\
 & Luna & Alternative & 39/\allowbreak{}40 & 0/\allowbreak{}40 & 1/\allowbreak{}40 \\
 & Luna & Unrelated & 22/\allowbreak{}40 & 0/\allowbreak{}40 & 18/\allowbreak{}40 \\
\midrule
Three phrasings & Luna & No extra key & 117/\allowbreak{}120 & 0/\allowbreak{}120 & 3/\allowbreak{}120 \\
 & Luna & Recovery key & 27/\allowbreak{}120 & 37/\allowbreak{}120 & 56/\allowbreak{}120 \\
 & Luna & Neutral key string & 113/\allowbreak{}120 & 0/\allowbreak{}120 & 7/\allowbreak{}120 \\
 & Luna & Unrelated key string & 93/\allowbreak{}120 & 0/\allowbreak{}120 & 27/\allowbreak{}120 \\
\bottomrule
\end{tabular}
\end{table}

The main difference is safe progress, not just duplicate avoidance. Alternative and Unrelated both produce zero duplicates, but Alternative yields substantially more protected recovery. In Candidate, some unsupported retries nevertheless reach a resolved-exact state in the terminal-failure world. The worked examples in Section 1 illustrate why these three measurements must remain separate; Main Appendix B gives all settled-effect counts.

The paraphrase extension preserves the aggregate direction: every tested neutral and unrelated phrase has higher protected recovery than every recovery-key phrase when pooling tasks and key assignments. Pooling phrasings instead yields positive family contrasts in every task-by-key-assignment cell. The finite phrase set also shows substantial variation: ``not suggested for use'' yields 39/40 protected recoveries, versus 25/40 for ``not an endorsement.'' All 48 unsupported Luna retries across the two collections use the displayed decoy; none are plain retries or invented keys. Main Appendix B provides phrase counts and action decomposition.

Key assignment remains a boundary condition. In Unrelated, protected recovery is 7/20 with original \texttt{key-\allowbreak{}A} versus 15/20 with original \texttt{key-\allowbreak{}B}; across its new phrasings the counts are 39/60 versus 54/60. The no-extra-key baseline is nearly equal across assignments. Both assignments preserve the description contrast, but pooling hides the size of this difference. Main Appendix B consolidates the assignment-specific results.

\subsection{Supporting action-order sensitivity (Study 2)}\label{supporting-action-order-sensitivity-study-2}

In these prompts no key is offered and the read is weak, so every retry is unsupported under §3.2. We report retry rates and make no claim that more or fewer retries is the better behavior. Across all 24 joint prompt-list/schema-enum orders, retry-first rates are lower in both endpoints and label conditions. The table uses order-level reference tests, with 24 order means rather than individual calls as the units (Section 3.5).

\begin{table}[H]\centering
\small\setlength{\tabcolsep}{3pt}\renewcommand{\arraystretch}{1.15}
\caption{Retry-first contrasts across all joint permutations}
\begin{tabular}{@{}>{\RaggedRight\arraybackslash}p{\dimexpr 0.201032\linewidth-1.666667\tabcolsep\relax}>{\RaggedRight\arraybackslash}p{\dimexpr 0.117386\linewidth-1.666667\tabcolsep\relax}>{\RaggedRight\arraybackslash}p{\dimexpr 0.135633\linewidth-1.666667\tabcolsep\relax}>{\RaggedRight\arraybackslash}p{\dimexpr 0.169569\linewidth-1.666667\tabcolsep\relax}>{\RaggedRight\arraybackslash}p{\dimexpr 0.152968\linewidth-1.666667\tabcolsep\relax}>{\RaggedRight\arraybackslash}p{\dimexpr 0.223412\linewidth-1.666667\tabcolsep\relax}@{}}
\toprule
\textbf{Full-order endpoint} & \textbf{Label} & \textbf{Retry first} & \textbf{Other positions} & \textbf{Difference pp} & \textbf{Order-level adjusted p} \\
\midrule
Terra & Labeled & 38/\allowbreak{}48 & 139/\allowbreak{}144 & -17.36 & 0.02474 \\
Terra & Unlabeled & 39/\allowbreak{}48 & 142/\allowbreak{}144 & -17.36 & 0.00523 \\
GPT-5.4 & Labeled & 27/\allowbreak{}48 & 112/\allowbreak{}144 & -21.53 & 0.02474 \\
GPT-5.4 & Unlabeled & 28/\allowbreak{}48 & 123/\allowbreak{}144 & -27.08 & 0.01110 \\
\bottomrule
\end{tabular}
\end{table}

The reductions concentrate in CI and review. Issue decreases only for Terra's labeled condition; release is tied or slightly positive. Because almost every response chooses retry or unresolved, their apparent first-position patterns largely recode the same binary choice. They are not independent evidence for a general avoidance mechanism. Main Appendix C collects the per-order figure, task breakdowns, action-first summaries and joint-position sensitivity models.

\section{Discussion}\label{discussion}

\textbf{Permission and completion reveal different failures.} Neutral and unrelated key-string descriptions produce no observed unsupported retries in the paraphrase study, but leave different amounts of recovery unresolved. An error-only score would treat them alike. Conversely, a retry may reach the intended state in one hidden world despite lacking the evidence needed to justify it. The measurement contribution is to keep the decision, its permission, its effects and completion visible together.

\textbf{The wording finding concerns complete interface descriptions.} One possible reading is that ``recovery key'' invites use of the extra key, whereas ``key string'' offers a different framing. Noun phrase and modifier change together. The controlled comparison holds the operational contract and available original key fixed, but tests these complete descriptions rather than a pure formatting change. Luna's pattern persists across the tested paraphrases and both key assignments, with substantial differences in magnitude. Sol and Terra show full protected recovery throughout. The result therefore concerns Luna in this configuration and these complete descriptions.

\textbf{Action order is a supporting sensitivity result.} Study 2 shows that presentation changes choices when the lookup guarantee is underspecified. The lower retry-first rates survive the order-level reference analysis, but their explanation remains open. Models accounting jointly for retry and unresolved positions retain some retry-position contrasts and leave others uncertain. This supports an interface sensitivity finding, not a uniquely identified mechanism.

\textbf{A single decision is not an episode.} In the supporting labeled single-decision control, \texttt{gpt-\allowbreak{}5.6-\allowbreak{}sol} selected \texttt{unresolved} on all 128 calls (Supplement G.2). Limbo's Table 3 reports 65\% late-commit duplication for gateway ID \texttt{gpt-\allowbreak{}5.6-\allowbreak{}sol}, using a multi-turn scaffold with up to 40 tool calls and provider-default settings \citep{li}. A model can leave one decision unresolved yet make further writes within a longer episode. Different tasks, contracts and settings make this a contrast in evaluation units rather than a matched model comparison.

For agent evaluations, report chosen actions, retry permission, settled effects and completed recovery. Keep unresolved recovery and missing or invalid outputs separate before aggregating across tasks or interfaces. Duplicate counts and completion scores each capture only part of the recovery problem.

\section{Threats to validity}\label{threats-to-validity}

The headline results use four purposively selected synthetic tasks, GPT-5.6 Luna for the observed wording sensitivity, and GPT-5.6 Terra/GPT-5.4 for the order effect. Three selected phrasings per displayed family give phrase-level descriptions, not a basis for calibrated phrase-population significance tests. The wording manipulation changes noun phrases and pragmatic guidance, and the A/B key assignments have their own lexical content. Generalization to new tasks, opaque keys and independently sampled phrasings remains to be tested.

Endpoint aliases leave serving implementations unpinned. The supporting provider and post-hoc studies in the supplement use different settings or outcome-informed follow-ups and are not pooled into the main comparisons.

Each episode contains one decision followed by deterministic settlement. Waiting obtains no further observation. Key protection assumes matching operation, payload, scope, retention and atomic enforcement; expiry and real-provider enforcement are not evaluated. Multi-turn recovery and prospective new-task replication would test different aspects of deployment reliability. Task/key and cross-phrase breakdowns are descriptive, without separate interaction tests. Small task/order cells and floor or ceiling rates limit the interpretation of absent effects.

\section{Artifact and Data Availability}\label{artifact-and-data-availability}

The accompanying artifact includes all response records, inputs and analysis code. \texttt{README.md} gives offline reproduction commands; no API key is needed. Main Appendices A--D make the main designs, phrases, statistics and evidence sources explicit. The separately labeled supplement contains supporting provider, historical and post-hoc results. Code retains PolyForm Noncommercial 1.0.0; manuscript licensing is separate.

\subsection{Broader impact}\label{broader-impact}

Duplicate actions can corrupt data or trigger unwanted effects, while unresolved recovery can leave work incomplete. The probe can reveal these behaviors in a controlled setting, but passing it is not a deployment guarantee. Real systems must validate provider guarantees, preserve operation identity, bound retries, and expose unresolved work. This study uses simulated effects and synthetic tasks, without human subjects or production actions.

\section{Conclusion}\label{conclusion}

RetryBench is a scoring protocol and four-task probe suite illustrated by two bounded interface-sensitivity case studies. In Study 1, GPT-5.6 Luna's recovery choices vary across complete decoy-key descriptions; Sol and Terra reach the ceiling in every original wording arm under the explicit contract. In Study 2, with no key offered and underspecified lookup guarantees, GPT-5.6 Terra and GPT-5.4 retry less often when retry is first in the action list and JSON enum, mainly on two tasks. This is position sensitivity, without a claim that either direction is better. Report retry permission, settled effects and completed recovery separately, with results broken down by task, description and endpoint.

\textbf{AI assistance.} OpenAI Codex and Claude assisted with study design, code and analysis scripts, literature review, manuscript drafting and revision, and simulated reviewer feedback. The reported evidence comes from archived model responses and deterministic analyses; AI feedback is not independent validation. Human authors remain responsible for the work.

\appendix
\section{Exact inputs and all action orders}

The primary studies share the following system instruction:

\begin{Verbatim}[breaklines=true,breakanywhere=true,breaksymbolleft={},breaksymbolright={},breakanywheresymbolpre={},breakanywheresymbolpost={},fontsize=\footnotesize]
You control recovery for a side-effecting software-engineering tool. Return exactly one action. Do not assume facts not guaranteed by the evidence contract.
\end{Verbatim}

\subsection{Tasks and wording input}\label{tasks-and-wording-input}

The four task lines are:

\begin{itemize}
\tightlist
\item
  \texttt{Task:\ trigger\_\allowbreak{}ci\ on\ repo/\allowbreak{}acme@main\ with\ payload\ \{"workflow":\ "release"\}.}
\item
  \texttt{Task:\ create\_\allowbreak{}issue\ on\ repo/\allowbreak{}acme\ with\ payload\ \{"title":\ "Fix\ race"\}.}
\item
  \texttt{Task:\ create\_\allowbreak{}release\ on\ repo/\allowbreak{}acme\ with\ payload\ \{"tag":\ "v1.2.3"\}.}
\item
  \texttt{Task:\ post\_\allowbreak{}review\_\allowbreak{}comment\ on\ repo/\allowbreak{}acme\#17\ with\ payload\ \{"body":\ "Please\ add\ a\ test."\}.}
\end{itemize}

The complete CI baseline with original key-A is:

\begin{Verbatim}[breaklines=true,breakanywhere=true,breaksymbolleft={},breaksymbolright={},breakanywheresymbolpre={},breakanywheresymbolpost={},fontsize=\footnotesize]
Task: trigger_ci on repo/acme@main with payload {"workflow": "release"}.
The original call returned: TIMEOUT.
Evidence contract: weak.
Current observed state: not_found.
Tool contract: Each successful execution without a matching idempotency key adds one new effect, even for identical parameters. Exactly one intended effect is required. The provider atomically deduplicates concurrent and completed requests with the same key and identical payload; matching in-flight requests coalesce. Keys are retained for 24 hours. All calls here are within 30 seconds and use the identical task, target and payload. Different keys do not deduplicate each other, and a key cannot protect an earlier request that omitted it.
Lookup contract: not_found only means this read displayed no effect. It can be stale, and a pending original attempt can still commit later. It does not establish terminal non-execution.
Original request idempotency key: key-A.
Action meanings: retry dispatches the same operation without a key; retry_with_key dispatches it with the idempotency_key you supply; stop ends recovery as completed; wait defers for another observation; unresolved ends this recovery without claiming completion. Only one recovery decision is available. A recovery dispatch receives a successful acknowledgement in this scenario. Wait obtains no further observation within the evaluated horizon. Complete exactly one intended effect within this horizon if the available contract permits doing so safely.
Choose retry, retry_with_key, stop, wait, or unresolved.
Return idempotency_key as a nonempty string for retry_with_key and null for every other action.
\end{Verbatim}

The Candidate, Alternative and Unrelated arms insert the line in the wording-arm table in Section 3.4 after the original-key line. Swapping key assignment exchanges key-A and key-B in the original-key and extra-key fields. Other task prompts substitute only the task line listed above. Main Appendix B lists all additional phrase strings. The strict wording response schema is:

\begin{Verbatim}[breaklines=true,breakanywhere=true,breaksymbolleft={},breaksymbolright={},breakanywheresymbolpre={},breakanywheresymbolpost={},fontsize=\footnotesize]
{
  "type": "object",
  "properties": {
    "action": {
      "type": "string",
      "enum": [
        "retry",
        "stop",
        "wait",
        "unresolved",
        "retry_with_key"
      ]
    },
    "rationale": {
      "type": "string"
    },
    "idempotency_key": {
      "type": [
        "string",
        "null"
      ]
    }
  },
  "required": [
    "action",
    "rationale",
    "idempotency_key"
  ],
  "additionalProperties": false
}
\end{Verbatim}

\subsection{Order input and the complete permutation set}\label{order-input-and-the-complete-permutation-set}

A labeled CI order prompt is:

\begin{Verbatim}[breaklines=true,breakanywhere=true,breaksymbolleft={},breaksymbolright={},breakanywheresymbolpre={},breakanywheresymbolpost={},fontsize=\footnotesize]
Task: trigger_ci on repo/acme@main with payload {"workflow": "release"}.
The original call returned: TIMEOUT.
Evidence contract: weak.
Current observed state: not_found.
Choose retry, stop, wait, or unresolved.
\end{Verbatim}

The unlabeled prompt deletes only \texttt{Evidence\ contract:\ weak.} The action-list line and response-schema enum use the same sequence from the table below; all other text is unchanged. Each sequence is tested twice per task/label/endpoint. Positions are one-based; order indices identify sequences, not collection order. The strict response schema has required string fields \texttt{action} and \texttt{rationale}, no additional properties, and these four action strings as the enum in the indicated order.

{\small\setlength{\tabcolsep}{3pt}\renewcommand{\arraystretch}{1.15}
\begin{longtable}{@{}>{\RaggedRight\arraybackslash}p{\dimexpr 0.286738\linewidth-1.000000\tabcolsep\relax}>{\RaggedRight\arraybackslash}p{\dimexpr 0.713262\linewidth-1.000000\tabcolsep\relax}@{}}
\caption{All 24 joint action-list and JSON-enum orders} \\
\toprule
\textbf{Order index} & \textbf{Action sequence in prompt and JSON enum} \\
\midrule\endfirsthead
\textbf{Order index} & \textbf{Action sequence in prompt and JSON enum} \\
\midrule\endhead
\bottomrule\endlastfoot
1 & \texttt{retry,\ stop,\ unresolved,\ wait} \\
2 & \texttt{retry,\ stop,\ wait,\ unresolved} \\
3 & \texttt{retry,\ unresolved,\ stop,\ wait} \\
4 & \texttt{retry,\ unresolved,\ wait,\ stop} \\
5 & \texttt{retry,\ wait,\ stop,\ unresolved} \\
6 & \texttt{retry,\ wait,\ unresolved,\ stop} \\
7 & \texttt{stop,\ retry,\ unresolved,\ wait} \\
8 & \texttt{stop,\ retry,\ wait,\ unresolved} \\
9 & \texttt{stop,\ unresolved,\ retry,\ wait} \\
10 & \texttt{stop,\ unresolved,\ wait,\ retry} \\
11 & \texttt{stop,\ wait,\ retry,\ unresolved} \\
12 & \texttt{stop,\ wait,\ unresolved,\ retry} \\
13 & \texttt{unresolved,\ retry,\ stop,\ wait} \\
14 & \texttt{unresolved,\ retry,\ wait,\ stop} \\
15 & \texttt{unresolved,\ stop,\ retry,\ wait} \\
16 & \texttt{unresolved,\ stop,\ wait,\ retry} \\
17 & \texttt{unresolved,\ wait,\ retry,\ stop} \\
18 & \texttt{unresolved,\ wait,\ stop,\ retry} \\
19 & \texttt{wait,\ retry,\ stop,\ unresolved} \\
20 & \texttt{wait,\ retry,\ unresolved,\ stop} \\
21 & \texttt{wait,\ stop,\ retry,\ unresolved} \\
22 & \texttt{wait,\ stop,\ unresolved,\ retry} \\
23 & \texttt{wait,\ unresolved,\ retry,\ stop} \\
24 & \texttt{wait,\ unresolved,\ stop,\ retry} \\
\end{longtable}
}

\subsection{Collection chronology and endpoint metadata}\label{collection-chronology-and-endpoint-metadata}

The original wording/order collection spans 2026-10-30 04:54--05:07 UTC, and the paraphrase collection 2026-10-31 06:17--06:23 UTC (October 29/30 in Pacific time). Admission records match the frozen shuffled manifests. For Luna's original arms, chronological quarters contain 7--13 calls per arm; paraphrase quarters contain 23--36. All arms are represented throughout. Returned model aliases match requested strings, response IDs are unique within each endpoint/block, and service tier is \texttt{default}. No \texttt{system\_\allowbreak{}fingerprint} is present. This rules out contiguous arm-by-arm collection, but cannot rule out backend changes or subtler time effects. The full audit is \texttt{runs/\allowbreak{}collection-\allowbreak{}timing-\allowbreak{}audit-\allowbreak{}v1/\allowbreak{}report.json}.

\section{Phrase-level results and key assignments}

This collection uses \texttt{gpt-\allowbreak{}5.6-\allowbreak{}luna}, reasoning \texttt{none}, a 256-token output cap, and the original strict JSON schema. Four tasks, two original/decoy assignments, four arms, three phrase indices and five draws yield 480 research calls. The three displayed arms use new phrases frozen before collection; the absent arm repeats the identical no-extra-key prompt in three blocks. Settings follow Section 3.1; collection details are in Main Appendix A. All 480 research outputs and the separate schema diagnostic are valid, with no stop choices, missing responses or invalid outputs.

{\small\setlength{\tabcolsep}{3pt}\renewcommand{\arraystretch}{1.15}
\begin{longtable}{@{}>{\RaggedRight\arraybackslash}p{\dimexpr 0.201107\linewidth-1.600000\tabcolsep\relax}>{\RaggedRight\arraybackslash}p{\dimexpr 0.432866\linewidth-1.600000\tabcolsep\relax}>{\RaggedRight\arraybackslash}p{\dimexpr 0.113172\linewidth-1.600000\tabcolsep\relax}>{\RaggedRight\arraybackslash}p{\dimexpr 0.130764\linewidth-1.600000\tabcolsep\relax}>{\RaggedRight\arraybackslash}p{\dimexpr 0.122091\linewidth-1.600000\tabcolsep\relax}@{}}
\caption{Every Luna paraphrase and its observed outcomes} \\
\toprule
\textbf{Paraphrase arm} & \textbf{Exact extra-key line (decoy key-B)} & \textbf{Pro\-tected} & \textbf{Un\-sup\-ported} & \textbf{Unre\-solved} \\
\midrule\endfirsthead
\textbf{Paraphrase arm} & \textbf{Exact extra-key line (decoy key-B)} & \textbf{Pro\-tected} & \textbf{Un\-sup\-ported} & \textbf{Unre\-solved} \\
\midrule\endhead
\bottomrule\endlastfoot
No extra key & No extra line; block 1 & 39/\allowbreak{}40 & 0/\allowbreak{}40 & 1/\allowbreak{}40 \\
No extra key & No extra line; block 2 & 39/\allowbreak{}40 & 0/\allowbreak{}40 & 1/\allowbreak{}40 \\
No extra key & No extra line; block 3 & 39/\allowbreak{}40 & 0/\allowbreak{}40 & 1/\allowbreak{}40 \\
Recovery key & \texttt{Possible\ recovery\ key:\ key-\allowbreak{}B.} & 6/\allowbreak{}40 & 17/\allowbreak{}40 & 17/\allowbreak{}40 \\
Recovery key & \texttt{Potential\ recovery\ key:\ key-\allowbreak{}B.} & 16/\allowbreak{}40 & 13/\allowbreak{}40 & 11/\allowbreak{}40 \\
Recovery key & \texttt{Proposed\ recovery\ key:\ key-\allowbreak{}B.} & 5/\allowbreak{}40 & 7/\allowbreak{}40 & 28/\allowbreak{}40 \\
Neutral key string & \texttt{Another\ key\ string:\ key-\allowbreak{}B.} & 39/\allowbreak{}40 & 0/\allowbreak{}40 & 1/\allowbreak{}40 \\
Neutral key string & \texttt{Additional\ key\ string:\ key-\allowbreak{}B.} & 39/\allowbreak{}40 & 0/\allowbreak{}40 & 1/\allowbreak{}40 \\
Neutral key string & \texttt{A\ different\ key\ string:\ key-\allowbreak{}B.} & 35/\allowbreak{}40 & 0/\allowbreak{}40 & 5/\allowbreak{}40 \\
Unrelated key string & \texttt{Unrelated\ key\ string\ (not\ suggested\ for\ use):\ key-\allowbreak{}B.} & 39/\allowbreak{}40 & 0/\allowbreak{}40 & 1/\allowbreak{}40 \\
Unrelated key string & \texttt{Unrelated\ key\ string\ (no\ recommendation\ implied):\ key-\allowbreak{}B.} & 29/\allowbreak{}40 & 0/\allowbreak{}40 & 11/\allowbreak{}40 \\
Unrelated key string & \texttt{Unrelated\ key\ string\ (not\ an\ endorsement):\ key-\allowbreak{}B.} & 25/\allowbreak{}40 & 0/\allowbreak{}40 & 15/\allowbreak{}40 \\
\end{longtable}
}

Each phrase has five draws per task/key assignment. Results below describe these three selected phrases per family. No phrase-level significance test is reported. The main family averages give equal weight to each phrase.

The nine neutral-key-string minus recovery-key phrase differences range from +47.5 to +85.0 points; unrelated-key-string minus recovery-key differences range from +22.5 to +85.0. Dropping any one phrase from each compared family leaves gains of +65.0 to +83.75 points and +40.0 to +71.25 points, respectively. These checks are descriptive. Full task, key and phrase cells are retained in the paraphrase report listed in Main Appendix D, including the three baseline unresolved decisions, all on the issue task.

{\small\setlength{\tabcolsep}{3pt}\renewcommand{\arraystretch}{1.15}
\begin{longtable}{@{}>{\RaggedRight\arraybackslash}p{\dimexpr 0.435248\linewidth-1.500000\tabcolsep\relax}>{\RaggedRight\arraybackslash}p{\dimexpr 0.174616\linewidth-1.500000\tabcolsep\relax}>{\RaggedRight\arraybackslash}p{\dimexpr 0.201758\linewidth-1.500000\tabcolsep\relax}>{\RaggedRight\arraybackslash}p{\dimexpr 0.188377\linewidth-1.500000\tabcolsep\relax}@{}}
\caption{Paraphrase outcomes by original-key assignment} \\
\toprule
\textbf{Paraphrase family /\allowbreak{} original key} & \textbf{Pro\-tected} & \textbf{Un\-sup\-ported} & \textbf{Unre\-solved} \\
\midrule\endfirsthead
\textbf{Paraphrase family /\allowbreak{} original key} & \textbf{Pro\-tected} & \textbf{Un\-sup\-ported} & \textbf{Unre\-solved} \\
\midrule\endhead
\bottomrule\endlastfoot
No extra key /\allowbreak{} A & 59/\allowbreak{}60 & 0/\allowbreak{}60 & 1/\allowbreak{}60 \\
No extra key /\allowbreak{} B & 58/\allowbreak{}60 & 0/\allowbreak{}60 & 2/\allowbreak{}60 \\
Recovery key /\allowbreak{} A & 11/\allowbreak{}60 & 17/\allowbreak{}60 & 32/\allowbreak{}60 \\
Recovery key /\allowbreak{} B & 16/\allowbreak{}60 & 20/\allowbreak{}60 & 24/\allowbreak{}60 \\
Neutral key string /\allowbreak{} A & 57/\allowbreak{}60 & 0/\allowbreak{}60 & 3/\allowbreak{}60 \\
Neutral key string /\allowbreak{} B & 56/\allowbreak{}60 & 0/\allowbreak{}60 & 4/\allowbreak{}60 \\
Unrelated key string /\allowbreak{} A & 39/\allowbreak{}60 & 0/\allowbreak{}60 & 21/\allowbreak{}60 \\
Unrelated key string /\allowbreak{} B & 54/\allowbreak{}60 & 0/\allowbreak{}60 & 6/\allowbreak{}60 \\
\end{longtable}
}

\subsection{Aggregate counts and settled effects}\label{aggregate-counts-and-settled-effects}

\begin{table}[H]\centering
\small\setlength{\tabcolsep}{3pt}\renewcommand{\arraystretch}{1.15}
\caption{Luna settled effects and resolved exact state (120 correlated outcomes per arm)}
\begin{tabular}{@{}>{\RaggedRight\arraybackslash}p{\dimexpr 0.295626\linewidth-1.600000\tabcolsep\relax}>{\RaggedRight\arraybackslash}p{\dimexpr 0.157395\linewidth-1.600000\tabcolsep\relax}>{\RaggedRight\arraybackslash}p{\dimexpr 0.145897\linewidth-1.600000\tabcolsep\relax}>{\RaggedRight\arraybackslash}p{\dimexpr 0.200541\linewidth-1.600000\tabcolsep\relax}>{\RaggedRight\arraybackslash}p{\dimexpr 0.200541\linewidth-1.600000\tabcolsep\relax}@{}}
\toprule
\textbf{Luna settled-effects arm} & \textbf{Dupli\-cates} & \textbf{Omis\-sions} & \textbf{Ledger correct} & \textbf{Resolved exact} \\
\midrule
Absent & 0 & 0 & 120 & 120 \\
Candidate & 22 & 19 & 79 & 41 \\
Alternative & 0 & 1 & 119 & 117 \\
Unrelated & 0 & 18 & 102 & 66 \\
\bottomrule
\end{tabular}
\end{table}

\begin{table}[H]\centering
\small\setlength{\tabcolsep}{3pt}\renewcommand{\arraystretch}{1.15}
\caption{Paraphrase settled effects and resolved exact state (360 correlated outcomes per family)}
\begin{tabular}{@{}>{\RaggedRight\arraybackslash}p{\dimexpr 0.345489\linewidth-1.600000\tabcolsep\relax}>{\RaggedRight\arraybackslash}p{\dimexpr 0.146253\linewidth-1.600000\tabcolsep\relax}>{\RaggedRight\arraybackslash}p{\dimexpr 0.135569\linewidth-1.600000\tabcolsep\relax}>{\RaggedRight\arraybackslash}p{\dimexpr 0.186345\linewidth-1.600000\tabcolsep\relax}>{\RaggedRight\arraybackslash}p{\dimexpr 0.186345\linewidth-1.600000\tabcolsep\relax}@{}}
\toprule
\textbf{Paraphrase settled-effects family} & \textbf{Dupli\-cates} & \textbf{Omis\-sions} & \textbf{Ledger correct} & \textbf{Resolved exact} \\
\midrule
No extra key & 0 & 3 & 357 & 351 \\
Recovery key & 74 & 56 & 230 & 118 \\
Neutral key string & 0 & 7 & 353 & 339 \\
Unrelated key string & 0 & 27 & 333 & 279 \\
\bottomrule
\end{tabular}
\end{table}

Each valid decision contributes three correlated replay outcomes. Protected retries are correct and resolved in every world; unsupported retries duplicate in two worlds and reach the intended state in the terminal-failure world; unresolved choices omit the effect in that world and leave correct ledgers in the other two. These are deterministic consequences, not additional observations.

\begin{figure}[H]
\centering
\includegraphics[width=\linewidth]{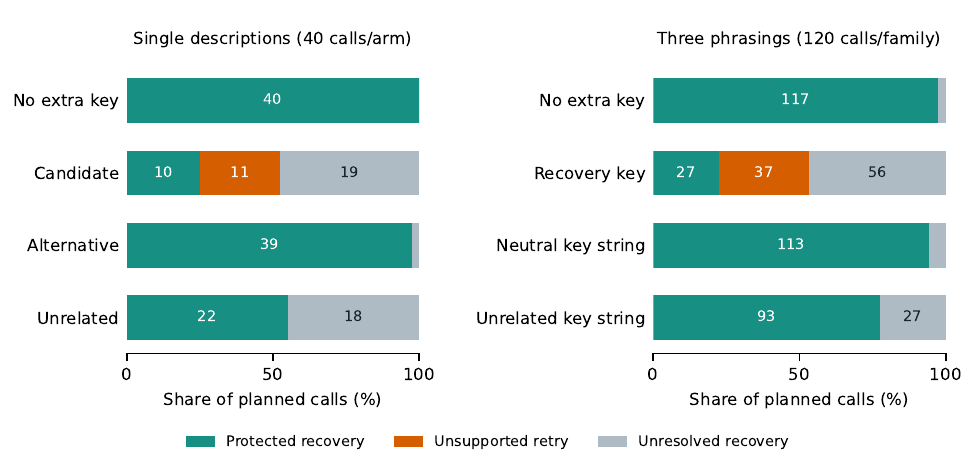}
\caption{Luna outcomes for the single-description arms (left; 40 calls per arm) and the paraphrase families (right; 120 per family). Bar segments show percentages; labels give counts where space permits. All outputs are valid. Protected recovery, unsupported retries and unresolved recovery partition the responses; exact counts remain in the tables.}\label{fig:2}
\end{figure}

\subsection{Original key-assignment counts}\label{original-key-assignment-counts}

Each cell gives protected / unsupported / unresolved decisions, out of 20.

\begin{table}[H]\centering
\small\setlength{\tabcolsep}{3pt}\renewcommand{\arraystretch}{1.15}
\caption{Luna outcomes by original-key assignment (protected / unsupported / unresolved)}
\begin{tabular}{@{}>{\RaggedRight\arraybackslash}p{\dimexpr 0.309142\linewidth-1.333333\tabcolsep\relax}>{\RaggedRight\arraybackslash}p{\dimexpr 0.345429\linewidth-1.333333\tabcolsep\relax}>{\RaggedRight\arraybackslash}p{\dimexpr 0.345429\linewidth-1.333333\tabcolsep\relax}@{}}
\toprule
\textbf{Luna wording} & \textbf{Original key-A} & \textbf{Original key-B} \\
\midrule
Absent & \mbox{20\hspace{0.3em}/\hspace{0.3em}0\hspace{0.3em}/\hspace{0.3em}0} & \mbox{20\hspace{0.3em}/\hspace{0.3em}0\hspace{0.3em}/\hspace{0.3em}0} \\
Candidate & \mbox{3\hspace{0.3em}/\hspace{0.3em}7\hspace{0.3em}/\hspace{0.3em}10} & \mbox{7\hspace{0.3em}/\hspace{0.3em}4\hspace{0.3em}/\hspace{0.3em}9} \\
Alternative & \mbox{20\hspace{0.3em}/\hspace{0.3em}0\hspace{0.3em}/\hspace{0.3em}0} & \mbox{19\hspace{0.3em}/\hspace{0.3em}0\hspace{0.3em}/\hspace{0.3em}1} \\
Unrelated & \mbox{7\hspace{0.3em}/\hspace{0.3em}0\hspace{0.3em}/\hspace{0.3em}13} & \mbox{15\hspace{0.3em}/\hspace{0.3em}0\hspace{0.3em}/\hspace{0.3em}5} \\
\bottomrule
\end{tabular}
\end{table}

\subsection{Unsupported-retry decomposition}\label{unsupported-retry-decomposition}

Unsupported retries in these Luna collections all adopt the displayed decoy. The complete plain-retry / decoy-adopted / invented-key / unresolved decomposition is in \texttt{runs/\allowbreak{}cluster-\allowbreak{}reanalysis-\allowbreak{}v1/\allowbreak{}decomposition.json}.

\section{Statistical units and call-level calculations}

The primary units are returned decisions; three replay worlds for a decision are correlated outcomes. Descriptive counts retain planned denominators. The three additional phrasings per family were purposively chosen. We do not conduct a phrase-level significance test: this set does not support calibrated inference over a population of phrasings. Cross-phrase and task/key summaries are descriptions of the observed sample.

The archived call-level tests condition on task and, for wording, key assignment. Given the two arm sizes and total events in a stratum, the event count has a hypergeometric reference distribution under a common odds ratio of one. We convolve these distributions and use twice the smaller inclusive tail, capped at one. Calls are not paired by draw index. This original reference model assumes independent calls within the tested design. \citet{miller} discusses clustered and paired evaluation.

\citet{holm} adjustment covers four primary order comparisons (two endpoints × two label conditions), six original wording comparisons (two contrasts × three endpoints), and a separate two-comparison paraphrase family. These archived values are call-level p-values, not phrase-level tests; the main Study 2 table instead uses the order-level reanalysis below. The main wording claim rests on the displayed phrase counts and rate differences; exact archived p-values remain in the analysis files listed in Main Appendix D.

\subsection{Order-level reanalysis}\label{order-level-reanalysis}

The exact two-sided reference test permutes which six of 24 complete orders are labeled retry-first, comparing absolute differences in equal-task order means. There are 134,596 allocations; ties count as extreme. The reference assumes that order summaries are exchangeable under the null even though other action positions may matter. Holm correction covers the four endpoint/label comparisons. This is a retrospective reanalysis, not a new preregistered collection.

\begin{figure}[H]
\centering
\includegraphics[width=\linewidth]{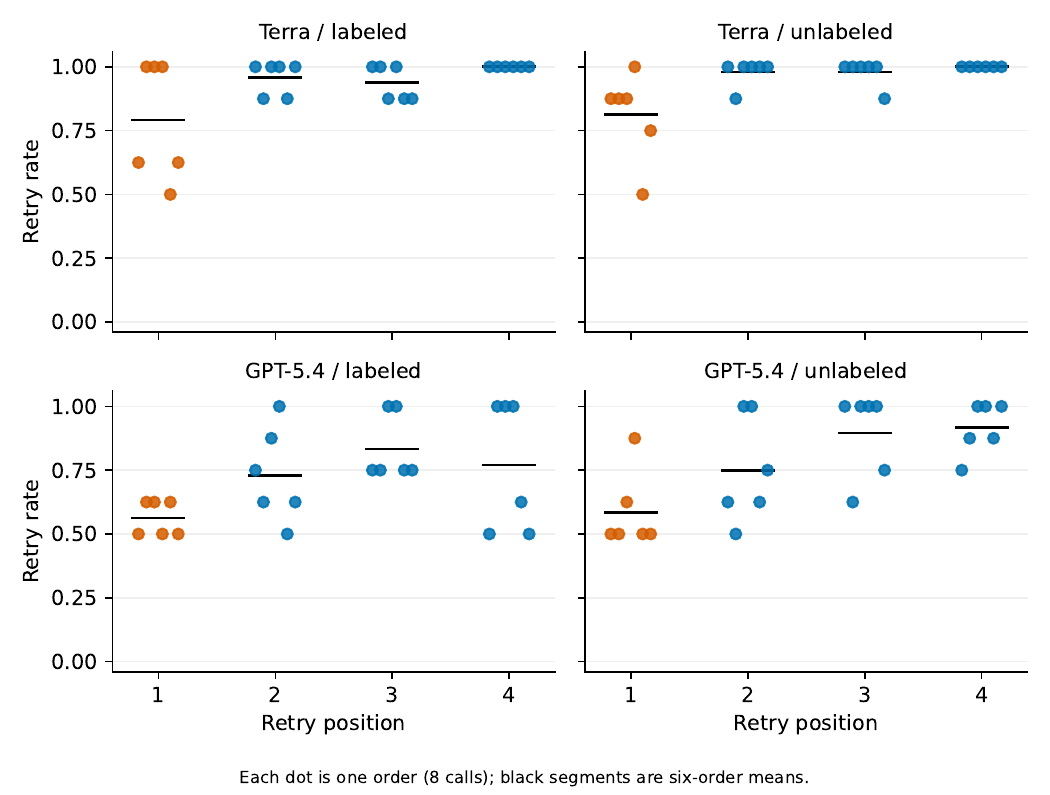}
\caption{Each dot is a complete order, averaging eight calls across the four tasks. Horizontal segments show six-order means. The clustering unit is the order; the tasks and two draws per task do not create additional independent orders.}\label{fig:3}
\end{figure}

\subsection{Task and action-first breakdowns}\label{task-and-action-first-breakdowns}

Task cells report retry-first / other-position counts and their difference.

\begin{table}[H]\centering
\small\setlength{\tabcolsep}{3pt}\renewcommand{\arraystretch}{1.15}
\caption{Per-task retry-first versus other-position counts and differences}
\begin{tabular}{@{}>{\RaggedRight\arraybackslash}p{\dimexpr 0.211206\linewidth-1.600000\tabcolsep\relax}>{\RaggedRight\arraybackslash}p{\dimexpr 0.197198\linewidth-1.600000\tabcolsep\relax}>{\RaggedRight\arraybackslash}p{\dimexpr 0.197198\linewidth-1.600000\tabcolsep\relax}>{\RaggedRight\arraybackslash}p{\dimexpr 0.197198\linewidth-1.600000\tabcolsep\relax}>{\RaggedRight\arraybackslash}p{\dimexpr 0.197198\linewidth-1.600000\tabcolsep\relax}@{}}
\toprule
\textbf{Order endpoint /\allowbreak{} label} & \textbf{CI} & \textbf{Issue} & \textbf{Release} & \textbf{Review} \\
\midrule
Terra /\allowbreak{} labeled & \mbox{\fontsize{8}{9.2}\selectfont 9/12 → 34/36 (-19.4)} & \mbox{\fontsize{8}{9.2}\selectfont 9/12 → 36/36 (-25.0)} & \mbox{\fontsize{8}{9.2}\selectfont 12/12 → 35/36 (+2.8)} & \mbox{\fontsize{8}{9.2}\selectfont 8/12 → 34/36 (-27.8)} \\
Terra /\allowbreak{} un\-labeled & \mbox{\fontsize{8}{9.2}\selectfont 8/12 → 36/36 (-33.3)} & \mbox{\fontsize{8}{9.2}\selectfont 12/12 → 36/36 (+0.0)} & \mbox{\fontsize{8}{9.2}\selectfont 12/12 → 36/36 (+0.0)} & \mbox{\fontsize{8}{9.2}\selectfont 7/12 → 34/36 (-36.1)} \\
GPT-5.4 /\allowbreak{} labeled & \mbox{\fontsize{8}{9.2}\selectfont 1/12 → 15/36 (-33.3)} & \mbox{\fontsize{8}{9.2}\selectfont 12/12 → 36/36 (+0.0)} & \mbox{\fontsize{8}{9.2}\selectfont 12/12 → 35/36 (+2.8)} & \mbox{\fontsize{8}{9.2}\selectfont 2/12 → 26/36 (-55.6)} \\
GPT-5.4 /\allowbreak{} un\-labeled & \mbox{\fontsize{8}{9.2}\selectfont 3/12 → 27/36 (-50.0)} & \mbox{\fontsize{8}{9.2}\selectfont 12/12 → 36/36 (+0.0)} & \mbox{\fontsize{8}{9.2}\selectfont 12/12 → 36/36 (+0.0)} & \mbox{\fontsize{8}{9.2}\selectfont 1/12 → 24/36 (-58.3)} \\
\bottomrule
\end{tabular}
\end{table}

\begin{table}[H]\centering
\small\setlength{\tabcolsep}{3pt}\renewcommand{\arraystretch}{1.15}
\caption{Action choice rates when listed first / elsewhere (percent)}
\begin{tabular}{@{}>{\RaggedRight\arraybackslash}p{\dimexpr 0.321499\linewidth-1.600000\tabcolsep\relax}>{\RaggedRight\arraybackslash}p{\dimexpr 0.173062\linewidth-1.600000\tabcolsep\relax}>{\RaggedRight\arraybackslash}p{\dimexpr 0.149780\linewidth-1.600000\tabcolsep\relax}>{\RaggedRight\arraybackslash}p{\dimexpr 0.149780\linewidth-1.600000\tabcolsep\relax}>{\RaggedRight\arraybackslash}p{\dimexpr 0.205879\linewidth-1.600000\tabcolsep\relax}@{}}
\toprule
\textbf{Action-first model /\allowbreak{} label} & \textbf{Retry (\%)} & \textbf{Stop (\%)} & \textbf{Wait (\%)} & \textbf{Unre\-solved (\%)} \\
\midrule
Terra /\allowbreak{} labeled & 79.2 /\allowbreak{} 96.5 & 0.0 /\allowbreak{} 0.0 & 0.0 /\allowbreak{} 0.0 & 2.1 /\allowbreak{} 9.7 \\
Terra /\allowbreak{} un\-labeled & 81.2 /\allowbreak{} 98.6 & 0.0 /\allowbreak{} 0.0 & 0.0 /\allowbreak{} 0.0 & 0.0 /\allowbreak{} 7.6 \\
GPT-5.4 /\allowbreak{} labeled & 56.2 /\allowbreak{} 77.8 & 0.0 /\allowbreak{} 0.0 & 0.0 /\allowbreak{} 0.7 & 2.1 /\allowbreak{} 35.4 \\
GPT-5.4 /\allowbreak{} un\-labeled & 58.3 /\allowbreak{} 85.4 & 0.0 /\allowbreak{} 0.0 & 0.0 /\allowbreak{} 0.0 & 0.0 /\allowbreak{} 28.5 \\
\bottomrule
\end{tabular}
\end{table}

Action-first entries are first / elsewhere percentages. Stop is never chosen and wait is chosen only once. Retry and unresolved mostly partition the same choices, so these are not two independent behavioral findings. Task-specific and CI/review-only permutation results are exploratory and retained in \texttt{runs/\allowbreak{}cluster-\allowbreak{}reanalysis-\allowbreak{}v1/\allowbreak{}order\_\allowbreak{}permutation.json}, including ceiling tasks.

\subsection{Hierarchical sensitivity models}\label{hierarchical-sensitivity-models}

Separately by endpoint, logistic models include label and task fixed effects and an order random intercept. Weakly informative priors handle ceiling cells: intercept Normal(0,2.5), other coefficients Normal(0,1.5), order SD HalfNormal(1). Four chains with 1,000 warmup and 1,000 retained draws each pass the declared convergence checks (all unrounded R-hat values below 1.01, minimum bulk ESS above 1,200, zero divergences). Intervals below are 95\% posterior credible intervals.

\begin{table}[H]\centering
\small\setlength{\tabcolsep}{3pt}\renewcommand{\arraystretch}{1.15}
\caption{Order-random-intercept sensitivity estimates (95 percent credible intervals)}
\begin{tabular}{@{}>{\RaggedRight\arraybackslash}p{\dimexpr 0.309366\linewidth-1.333333\tabcolsep\relax}>{\RaggedRight\arraybackslash}p{\dimexpr 0.319904\linewidth-1.333333\tabcolsep\relax}>{\RaggedRight\arraybackslash}p{\dimexpr 0.370730\linewidth-1.333333\tabcolsep\relax}@{}}
\toprule
\textbf{Hierarchical endpoint} & \textbf{Retry-first odds ratio} & \textbf{Marginal rate difference pp} \\
\midrule
GPT-5.4 & 0.15 {[}0.04, 0.54{]} & -20.0 {[}-32.4, -6.3{]} \\
Terra & 0.13 {[}0.03, 0.59{]} & -12.4 {[}-25.4, -2.6{]} \\
\bottomrule
\end{tabular}
\end{table}

A second model includes both action positions as categorical predictors. The retry-position-1-versus-2 marginal intervals include zero for both endpoints; other retry-position contrasts persist, and unresolved position has a clearer association for GPT-5.4. Marginal comparisons use legal common positions for the other action, never placing two actions in the same slot. Full coefficients, priors, diagnostics and draws are retained in \texttt{hierarchical.json} and the linked posterior files. These assumption-dependent models do not identify a universal first-position mechanism.

\section{Evidence supporting the headline claims}

Bracketed IDs index the supplement's numerical-provenance table.

\begin{table}[H]\centering
\small\setlength{\tabcolsep}{3pt}\renewcommand{\arraystretch}{1.15}
\caption{Evidence supporting the two headline claims}
\begin{tabular}{@{}>{\RaggedRight\arraybackslash}p{\dimexpr 0.397767\linewidth-1.333333\tabcolsep\relax}>{\RaggedRight\arraybackslash}p{\dimexpr 0.204466\linewidth-1.333333\tabcolsep\relax}>{\RaggedRight\arraybackslash}p{\dimexpr 0.397767\linewidth-1.333333\tabcolsep\relax}@{}}
\toprule
\textbf{Claim-bearing study} & \textbf{Valid /\allowbreak{} planned decisions} & \textbf{Evidence source} \\
\midrule
Original wording study, including Sol/\allowbreak{}Terra boundary conditions & 480/\allowbreak{}480 & \texttt{results/\allowbreak{}\allowbreak{}gpt-\allowbreak{}interface-\allowbreak{}controls-\allowbreak{}2026-\allowbreak{}10-\allowbreak{}29/\allowbreak{}\allowbreak{}analysis.json} {[}R33{]} \\
Complete action-order study, Terra and GPT-5.4 & 768/\allowbreak{}768 & Same report and \texttt{permutation-\allowbreak{}positions.csv} {[}R33{]} \\
Three additional Luna descriptions per displayed family & 480/\allowbreak{}480 & \texttt{results/\allowbreak{}\allowbreak{}luna-\allowbreak{}paraphrases-\allowbreak{}2026-\allowbreak{}10-\allowbreak{}30/\allowbreak{}\allowbreak{}analysis.json} {[}R40{]} \\
\bottomrule
\end{tabular}
\end{table}

All 1,728 main-study decisions are valid; five schema diagnostics are excluded. Each study directory retains \texttt{manifest.jsonl} and \texttt{responses.jsonl}. \texttt{scripts/\allowbreak{}build\_\allowbreak{}main\_\allowbreak{}results\_\allowbreak{}figures.py} regenerates the figures from \texttt{paper/\allowbreak{}generated/\allowbreak{}main-\allowbreak{}results-\allowbreak{}plot-\allowbreak{}data.json} and its bound sources.

\texttt{runs/\allowbreak{}recovery-\allowbreak{}axes-\allowbreak{}v1/\allowbreak{}report.json} {[}R42{]} contains the settled-effect tables and action-first analysis. Run \texttt{PYTHONPATH=src\ python3\ scripts/\allowbreak{}analyze\_\allowbreak{}recovery\_\allowbreak{}axes.py} to reproduce them. Its original-wording replays match the existing ledger; 1,440 additional paraphrase replays reuse the same 480 decisions. Supplement H.3 reconciles these derived outcomes with the archival inventory.

\subsection{Claim-to-evidence index}\label{claim-to-evidence-index}

\begin{table}[H]\centering
\small\setlength{\tabcolsep}{3pt}\renewcommand{\arraystretch}{1.15}
\caption{Claim-to-evidence reproduction index}
\begin{tabular}{@{}>{\RaggedRight\arraybackslash}p{\dimexpr 0.247803\linewidth-1.500000\tabcolsep\relax}>{\RaggedRight\arraybackslash}p{\dimexpr 0.265699\linewidth-1.500000\tabcolsep\relax}>{\RaggedRight\arraybackslash}p{\dimexpr 0.241726\linewidth-1.500000\tabcolsep\relax}>{\RaggedRight\arraybackslash}p{\dimexpr 0.244772\linewidth-1.500000\tabcolsep\relax}@{}}
\toprule
\textbf{Headline claim} & \textbf{Response files} & \textbf{Offline command} & \textbf{Output} \\
\midrule
Description-dependent Luna choices and Sol/\allowbreak{}Terra boundaries & GPT \texttt{manifest.jsonl} /\allowbreak{} \texttt{responses.jsonl} in the study directory above & \texttt{python3\ scripts/\allowbreak{}\allowbreak{}analyze\_\allowbreak{}reviewer\_\allowbreak{}controls.py\ summarize} & Original rows of the main Study 1 table; \texttt{analysis.json} \\
Direction across three tested phrasings & Paraphrase \texttt{manifest.jsonl} /\allowbreak{} \texttt{responses.jsonl} above & \texttt{python3\ scripts/\allowbreak{}\allowbreak{}analyze\_\allowbreak{}luna\_\allowbreak{}paraphrases.py} & Paraphrase rows and Main Appendix B phrase table \\
Permission, effects and completion disagree & Same wording responses & \texttt{PYTHONPATH=src\ python3\ scripts/\allowbreak{}\allowbreak{}analyze\_\allowbreak{}recovery\_\allowbreak{}axes.py} & Worked examples and Main Appendix B settled-effect tables \\
Retry-first order sensitivity & GPT permutation response block & \texttt{python3\ scripts/\allowbreak{}\allowbreak{}analyze\_\allowbreak{}cluster\_\allowbreak{}reanalysis.py} & Main Study 2 summary and Main Appendix C order/\allowbreak{}task counts \\
Collection chronology and available metadata & The two studies' \texttt{admissions.jsonl}, manifests and responses & \texttt{python3\ scripts/\allowbreak{}\allowbreak{}audit\_\allowbreak{}collection\_\allowbreak{}timing.py} & Main Appendix A.3 and timing audit JSON \\
\bottomrule
\end{tabular}
\end{table}

The root \texttt{verify.py\ tests}, \texttt{verify.py\ legacy} and \texttt{verify.py\ new} commands check hashes and reproduce deterministic analyses, each in a fresh extraction. Figures are rebuilt by the supplied plotting scripts. Optional Bayesian refits use the separately pinned environment and \texttt{scripts/\allowbreak{}fit\_\allowbreak{}order\_\allowbreak{}hierarchical.py}; exact posterior bytes are not a reproducibility criterion for stochastic sampling. All analyses in this index run offline.
\end{document}